\documentclass[runningheads]{llncs}
\usepackage[T1]{fontenc}
\usepackage{graphicx}
\usepackage{utfsym}
\usepackage{multirow}
\usepackage{marvosym}
\usepackage{amsmath}
\usepackage{hyperref}
\begin{document}
\title{Surgical-MambaLLM: Mamba2-enhanced Multimodal Large Language Model for VQLA in Robotic Surgery}
\author{Pengfei Hao\inst{1} \and
Hongqiu Wang\inst{1} \and
Shuaibo Li\inst{1} \and
Zhaohu Xing\inst{1}\and
Guang Yang\inst{2}\and
Kaishun Wu \inst{1} \and
Lei Zhu\inst{1,3}\textsuperscript{(\Letter)}}
\institute{Hong Kong University of Science and Technology (Guangzhou), Guangzhou, China \email{leizhu@ust.hk} \and
Imperial College London \\
\and 
The Hong Kong University of Science and Technology\\
}
%
%\titlerunning{Abbreviated paper title}
% If the paper title is too long for the running head, you can set
% an abbreviated paper title here
%
\if 0
\author{First Author\inst{1}\orcidID{0000-1111-2222-3333} \and
Second Author\inst{2,3}\orcidID{1111-2222-3333-4444} \and
Third Author\inst{3}\orcidID{2222--3333-4444-5555}}
%
\authorrunning{F. Author et al.}
% First names are abbreviated in the running head.
% If there are more than two authors, 'et al.' is used.
%
\institute{Princeton University, Princeton NJ 08544, USA \and
Springer Heidelberg, Tiergartenstr. 17, 69121 Heidelberg, Germany
\email{lncs@springer.com}\\
\url{http://www.springer.com/gp/computer-science/lncs} \and
ABC Institute, Rupert-Karls-University Heidelberg, Heidelberg, Germany\\
\email{\{abc,lncs\}@uni-heidelberg.de}}
\fi
%
\maketitle % typeset the header of the contribution
\begin{abstract}
In recent years, Visual Question Localized-Answering in robotic surgery (Surgical-VQLA) has gained significant attention for its potential to assist medical students and junior doctors in understanding surgical scenes. Recently, the rapid development of Large Language Models (LLMs) has provided more promising solutions for this task. However, current methods struggle to establish complex dependencies between text and visual details, and have difficulty perceiving the spatial information of surgical scenes. To address these challenges, we propose a novel method, Surgical-MambaLLM, which is the first to combine Mamba2 with LLM in the surgical domain, that leverages Mamba2's ability to effectively capture cross-modal dependencies and perceive spatial information in surgical scenes, thereby enhancing the LLMs' understanding of surgical images. Specifically, we propose the Cross-modal Bidirectional Mamba2 Integration (CBMI) module to leverage Mamba2 for effective multimodal fusion, with its cross-modal integration capabilities. Additionally, tailored to the geometric characteristics of surgical scenes, we design the Surgical Instrument Perception (SIP) scanning mode for Mamba2 to scan the surgical images, enhancing the model's spatial understanding of the surgical scene. Extensive experiments demonstrate that our Surgical-MambaLLM model outperforms the state-of-the-art methods on the EndoVis17-VQLA and EndoVis18-VQLA datasets, significantly improving the performance of the Surgical-VQLA task. 

\keywords{Robotic-assisted surgery  \and Multimodal Large Language Model \and Mamba \and Surgical visual question localized-answering}
\end{abstract}
\section{Introduction}
Intelligent robotic surgery systems and the task of Visual Question Localized-Answering in robotic surgery (Surgical-VQLA) have garnered significant attention in recent years \cite{b3}. Surgical-VQLA can answer questions about organs, the location of surgical instruments, and surgical procedures based on surgical images, while simultaneously providing the corresponding bounding boxes. This task not only assists medical students and junior residents in resolving questions about surgical procedures, thereby reducing the workload of experienced surgeons \cite{b1}, but also enhances the interpretative capabilities of intelligent surgical robotic systems in understanding surgical scenes \cite{b3,wang2024video,wang2023dynamic}.

%Most current Surgical-VQLA methods \cite{b3,b6,b7} use pre-trained vision and language models as backbone networks and have achieved significant success in performance. %In recent years, Large Language Models (LLMs) have made remarkable progress in natural language processing (NLP) and reasoning, and with the advent of Multimodal Large Language Models (MLLMs), they can now also process and understand visual scenes and multimodal information \cite{openai2023gpt,touvron2023llama, b5}. Inspired by this, \cite{b4} first introduced LLMs into the surgical domain and achieved remarkable results, illustrating the immense potential of LLMs in this field. 

%这里要重写motivation
Current Surgical-VQLA methods \cite{b3,b6,b7,hao2025enhancing} have achieved significant success in performance by using pre-trained vision and language models as backbone networks. In recent years, Large Language Models (LLMs) have made remarkable progress in natural language processing (NLP) \cite{openai2023gpt}. Inspired by this, \cite{b4} first introduced LLMs into the surgical domain and achieved remarkable results, illustrating the immense potential of LLMs in this field. However, these methods still have some limitations. Firstly, current methods primarily rely on the Transformer-based method \cite{vaswani2017attention,li2024unionformer} for cross-modal fusion, which causes the models to focus more on global features while neglecting local details. Thus, it becomes challenging to capture visual details and establish dependencies with the text, which are essential for accurately answering questions related to the state of the instruments. Additionally, although some research \cite{b4,hao2025enhancing,hao2025surgery} has introduced LLMs to surgery-related tasks, LLMs still face significant challenges in understanding surgical scenes, particularly in perceiving spatial information due to the complexity of laparoscopic environments.

Recently, the Mamba model, leveraging its State Space Models (SSMs) \cite{b21}, has demonstrated the ability to efficiently capture complex dependencies within sequences while preserving detailed information. The original Mamba model performs unidirectional scanning of text sequences, which is not suitable for spatially-aware visual tasks \cite{b18}. However, some methods \cite{b18,b19,wu2024rainmamba} have proved that multi-directional scanning can enhance Mamba's spatial perception capabilities for 2D vision. Inspired by this, we explore the possibility of combining Mamba with LLMs to leverage Mamba's ability to effectively capture cross-modal dependencies and perceive spatial information in surgical scenes, thereby enhancing LLMs' understanding of surgical images.

In this work, we propose the Surgical-MambaLLM model, the first method in the surgical domain to combine the Mamba2 \cite{b23} model with LLM to address Surgical-VQLA. To achieve effective cross-modal fusion and accurately perceive spatial information in surgical scenes, we design the Cross-modal Bidirectional Mamba2 Integration (CBMI) module. This module facilitates the establishment of complex dependencies between visual details and questions. Within CBMI, the Surgical Instrument Perception (SIP) scanning mode is designed instead of unidirectional sequence scanning \cite{b16}, which can enhance Mamba2's spatial awareness and understanding of surgical images. Subsequently, the fused features are processed through a projector, concatenated with the textual features, and then input into the LLM. To improve the performance of the LLM, we fine-tune it using the LoRA \cite{b24} technique, thereby obtaining the final feature representation. Finally, the output features are fed into the answer prediction head and the position prediction head to obtain the final prediction results.

Overall, contributions are as follows: (1) We propose the Surgical-MambaLLM, which is the first method that integrates Mamba2 with Large Language Model in the surgical domain. By leveraging Mamba2's ability to capture complex dependencies between different modalities, we achieve effective multimodal fusion, significantly enhancing LLM performance in Surgical-VQLA tasks. (2) We introduce the Cross-modal Bidirectional Mamba2 Integration (CBMI) module, which explores effective methods for fusing visual and textual data within Mamba2. (3) We design an innovative Surgical Instrument Perception (SIP) scanning mode in CBMI to enhance Mamba2's spatial understanding of surgical images. (4) We conduct extensive experiments to validate the effectiveness of Surgical-MambaLLM in the Surgical-VQLA task. The experimental results demonstrate that our approach outperforms other State-Of-The-Art (SOTA) models on publicly available EndoVis17-VQLA and EndoVis18-VQLA datasets.

\begin{figure*}[t]
    \centering
    \includegraphics[width=1\linewidth]{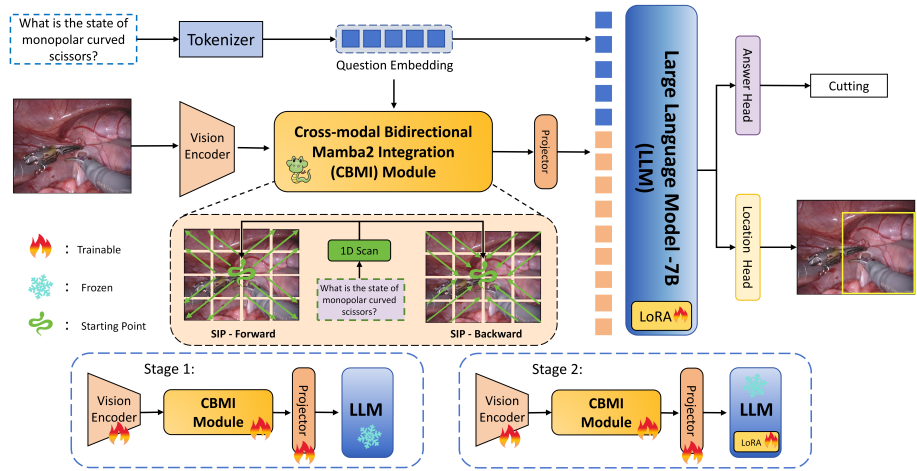}

    \caption{An overview of the framework of our Surgical-MambaLLM. Questions are input into the tokenizer to obtain the question embedding, while surgical images are processed by the vision encoder to extract the visual features. These features are integrated within the CBMI module, which utilizes our SIP scanning mode to scan the vision features and employs modified bidirectional Mamba2 blocks for multimodal feature fusion. The fused features are then projected into the LLM to generate answer and location predictions. The training process involves two stages: initially training the vision encoder, CBMI, and projector with frozen LLM parameters, followed by fine-tuning LLM using LoRA.}
    \label{fig2}

\end{figure*}

%It takes video frames and corresponding text descriptions as input, and segments the referred objects.
\section{Methodology}
%%%没写完
We propose the Surgical-MambaLLM model, a novel method integrating Mamba2 with LLM to improve LLM’s understanding of surgical scenes. The overall architecture of our Surgical-MambaLLM is illustrated in Fig.~\ref{fig2}.

\begin{figure}[t]
    \centering
    \includegraphics[width=1\linewidth]{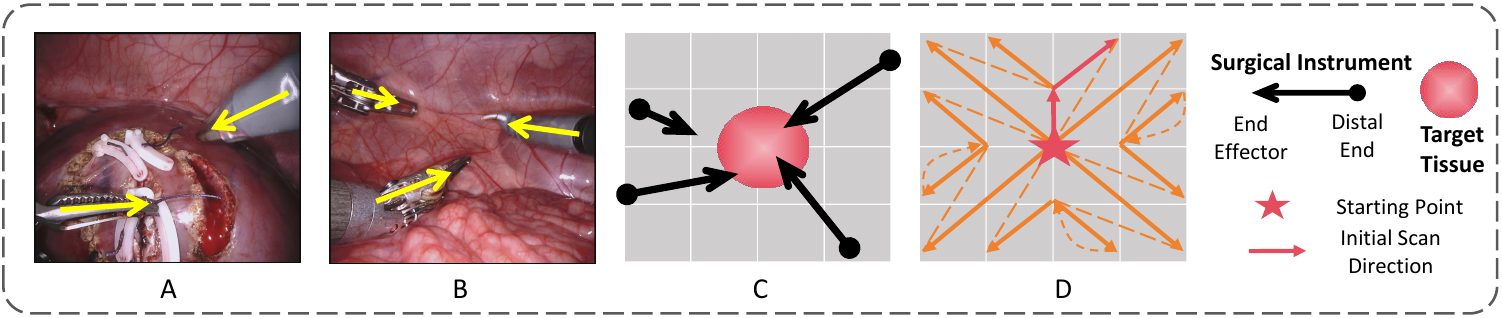}    
    \caption{A and B illustrate the directions of surgical instrument operations in surgical images; C represents the geometric modeling of the surgical scene; D is the Surgical Instrument Perception (SIP) scanning mode we proposed.}
    \label{fig1} 
    
\end{figure}

\subsection{Cross-modal Bidirectional Mamba2 Integration Module}
\subsubsection{Surgical Instrument Perception Scanning Mode}
In robotic surgery, surgeons typically operate from a console, observing the procedure on screens and manipulating control sticks to guide the robotic instruments \cite{b9}. To facilitate the surgeon's operation, the target organ or tissue is usually in the center of the laparoscopic image, while the ends of the robotic instruments extend from the periphery towards the central target organ \cite{b10}, as illustrated in Fig.~\ref{fig1} A and Fig.~\ref{fig1} B. This observation indicates that the operation of surgical instruments in laparoscopic surgery exhibits a clear directionality from the periphery towards the center, which can be geometrically abstracted as a radial pattern from the center outward, as shown in Fig.~\ref{fig1} C. Therefore, we propose the Surgical Instrument Perception (SIP) scanning mode for the Mamba2 model, which performs a radial scan from the center towards four directions, ultimately scanning the entire image to obtain a global representation, as shown in Fig.~\ref{fig1} D. By scanning from the center outward, the SIP scanning mode is highly likely to capture continuous regions of instrument features within the image, rather than dispersing the instrument's image features across different scanning areas. This approach helps maintain the integrity of the instrument regions, thereby enhancing the mamba2's spatial awareness of the surgical scene.

%We design the Surgical Instrument Perception (SIP) scanning mode, introducing an innovative center-to-periphery diagonal scanning method tailored for laparoscopic surgical robot vision. This approach captures global features by scanning from the image center outward, effectively capturing surgical details through the perception of instrument motion and operational nuances. By dividing the image into four regions, this method facilitates spatial analysis, enabling a better understanding of instrument activities and improving responses to location-based queries.

\begin{figure}[t]
    \centering
    \includegraphics[width=0.8\linewidth]{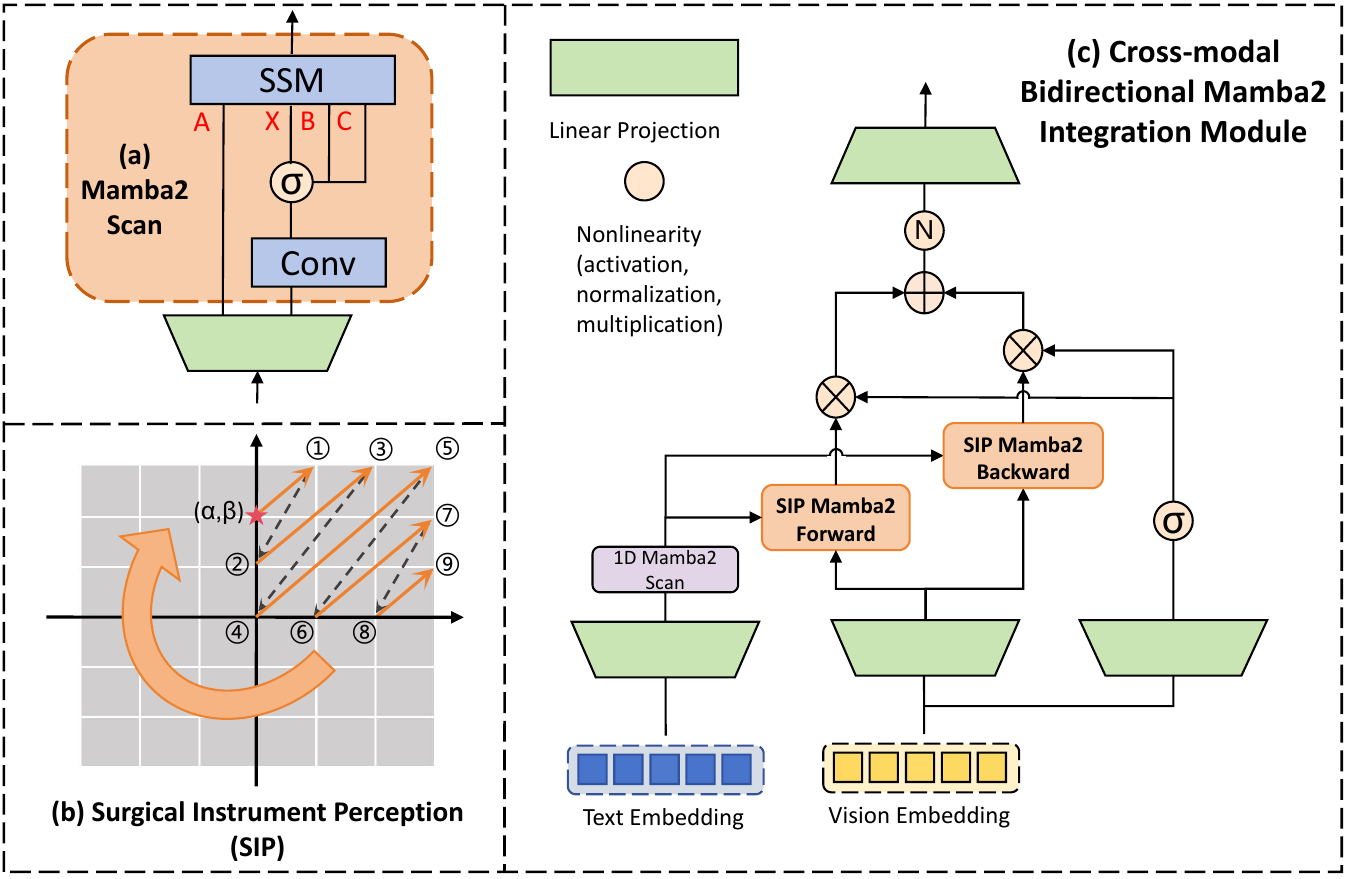}
    
    \caption{(a) is the architecture of Mamba2 Scan; (b) presents the details of the scanning track; (c) is the framework of the CBMI module.}
    \label{fig3}
    
\end{figure}

Specifically, taking the first quadrant as an example, the scanning starts at \((\alpha, \beta)\), as shown in Fig.~\ref{fig3} (b). The maximal value of $x_n$ or $y_n$ is \(N\), and the trajectory can be described by the following formula:
\begin{equation}
    (x_{n+1}, y_{n+1}) = 
\begin{cases} 
(0, y_n - k_n) & \text{if } y_n = N, x_n \neq N\\
(x_n - k_n, 0) & \text{if } x_n  = N \\
(x_n +1, y_n +1)
\end{cases} ,
\end{equation}
\begin{equation}
    k_n = 
\begin{cases} 
x_n + 1 & \text{if }y_n > x_n \\ 
y_n - 1 & \text{if } y_n \le x_n 
\end{cases} ,
\end{equation}
where $(x_n, y_n)$ is the current point, $(x_n +1, y_n +1)$ is the next point and the initial point is \((x_0, y_0) = (\alpha, \beta)\). $k_n$ is a dependent variable that can be denoted as (2). Subsequently, we sequentially scan the fourth, third, and second quadrants, ultimately obtaining the features of the entire image.

\subsubsection{Cross-modal Bidirectional Mamba2}
In the multimodal domain, most existing Mamba methods \cite{b27,zhao2024cobra} typically only concatenate visual features with textual features without designing fusion methods specifically for the two modalities. Therefore, we propose the Cross-modal Bidirectional Mamba2, which performs bidirectional scanning of visual features and textual features through the SIP scanning mode to achieve efficient feature fusion and improve the model's spatial understanding of surgical scenes, as shown in Fig.~\ref{fig3} (c).

Specifically, first, the text embedding \(t\) and the visual embedding \(v\) are preliminarily processed through linear projections to get visual feature $F_t$ and text feature $F_v$:
\begin{equation}
    F_t = l_t(t),  F_v = l_v(v), 
\end{equation}
where \(l_t\) and \(l_{v}\) are linear layer for feature projection.  

Subsequently, \(F_v\) and \(F_t\) are input into the SIP-Mamba2 Forward and SIP-Mamba2 Backward module. In this module, \(F_t\) undergoes 1D scanning, and \(F_v\) undergoes bidirectional scanning in the SIP scanning mode to obtain sequence features in opposite directions. Then \(F_t\) and the bidirectional \(F_v\) are input into the Mamba2 Scan to obtain the fused forward sequence features \(S_{\text{forward}}\) and backward sequence features \(S_{\text{backward}}\). The structure of the Mamba2 Scan is shown in Fig.~\ref{fig3} (a). The formula is as follows:
\begin{equation}
S_{\text{forward}} = \text{SIP-Mamba2}_{\text{forward}}(F_t, F_v), 
\end{equation}
\begin{equation}
S_{\text{backward}} = \text{SIP-Mamba2}_{\text{backward}}(F_t, F_v),  
\end{equation}
where $\text{SIP-Mamba2}_{\text{forward}}$ is SIP-Mamba2 Forward module, and $\text{SIP-Mamba2}_{\text{backward}}$ is SIP-Mamba2 Backward module. Next, \(S_{\text{forward}}\) and \(S_{\text{backward}}\) are multiplied by the original features \(F_v\) activated by \(\sigma\) and then added together to obtain the fused features \(S\). Finally, normalization and linear projection are applied to obtain the output features \(S_{\text{output}}\). The formula is as follows:
\begin{equation}
    S = S_{\text{forward}} \cdot \sigma(F_v) + S_{\text{backward}} \cdot \sigma(F_v),
\end{equation}
\begin{equation}
   S_{\text{output}} = Linear(LN(S)),
\end{equation}
where $\sigma$ is the activation function, $Linear$ is a linear layer and $LN$ denotes Layer normalization.

\subsection{Training Strategy}
Our training strategy consists of two stages for efficient model training. In the first stage, we freeze the parameters of LLM and train the vision encoder, CBMI module, projector, answer head and location head. This stage aims to enable the CBMI module to fuse visual and textual features effectively and adapt its output to the LLM via the projector. This process ensures that the vision encoder and the CBMI module can fully understand and process the complex information present in surgical scenarios. In the second stage, we fine-tune the LLM using LoRA techniques, applying a lower learning rate. This fine-tuning further optimizes the LLM for surgical-VQLA tasks while preserving the initial learning outcomes.

\section{Experiments and Results}

\subsection{Dataset} 
\textbf{EndoVis-18-VQLA}
% The EndoVis-18 Dataset, originating from the MICCAI Endoscopic Vision Challenge 2018 \cite{b11}, is a publicly available dataset comprising video sequences from 14 robotic surgical procedures. The annotations for EndoVis-18-VQLA are publicly accessible via \cite{b3}. Recent work \cite{b7} has expanded the training set from 9014 to 12741 QA pairs and the test set from 2769 to 3820 QA pairs. To enhance the robustness, credibility, and fairness of our study, we utilized this expanded dataset. The training set consists of 1560 frames, while the test set includes 436 frames. Following the guidelines in \cite{b3,b6,b7}, the dataset is split to ensure proper isolation between video sequences and prevent potential data leakage, with Training Video IDs: 2, 3, 4, 6, 7, 9, 10, 11, 12, 14, 15, and Test Video IDs: 1, 5, 16
This Dataset from the 2018 MICCAI Endoscopic Vision Challenge \cite{b11} includes video sequences from 14 surgeries. The annotations for EndoVis-18-VQLA are accessible via \cite{b3}. Recent work \cite{b7} has expanded the training set from 9014 to 12741 QA pairs and the test set from 2769 to 3820 QA pairs. Following \cite{b7}, we used the expanded dataset to ensure robustness and credibility in our study. Following previous works \cite{b3,b6,b7}, it includes 1,560 training frames and 436 test frames.\\
\textbf{EndoVis-17-VQLA}
% The EndoVis-17-VQLA Dataset originates from the MICCAI Endoscopic Vision Challenge in 2017 \cite{b12}. The original dataset includes 10 video sequences that document the process of robotic surgeries. The annotations for the EndoVis-17-VQLA Dataset can be publicly accessed via the referenced literature \cite{b3}. Following \cite{b3,b6,b7} to validate the generalization ability of our model on unseen data, we use EndoVis-17-VQLA as an external validation dataset. \cite{b7} also extend QA pairs from 472 to 708. To enhance the robustness, credibility, and fairness of our study, we utilized this expanded dataset. Our model will be evaluated on this dataset to demonstrate its effectiveness and robustness when dealing with new, unseen surgical scenarios.
This Dataset from the 2017 MICCAI Endoscopic Vision Challenge \cite{b12} features 10 robotic surgery video sequences. Annotations are available publicly \cite{b3}. Following guidelines in \cite{b3,b6,b7}, we use this dataset for external validation to test our model's generalization on unseen data. The dataset has been expanded from 472 to 708 QA pairs \cite{b7}, enhancing our study's robustness, credibility, and fairness. Models' performance will be evaluated on these new, unseen surgical scenarios.
\begin{table*}[t]
\centering
\caption{Comparison experiments between our Surgical-MambaLLM and other methods on EndoVis-18 and EndoVis-17 datasets}

\label{tab:1}
\resizebox{1\textwidth}{!}{%
\begin{tabular}{c|ccc|ccc}
\hline
\multirow{2}{*}{Models} & \multicolumn{3}{c|}{EndoVis - 18}          & \multicolumn{3}{c}{EndoVis - 17}                    \\ \cline{2-7} 
& Acc & F-Score & mIoU & Acc & F-Score & mIoU  \\ \hline \hline
VisualBERT \cite{b1}    & 0.6234  & 0.3269 & 0.7336   & 0.4516  & 0.2698   & 0.7268          \\
VisualBERT RM \cite{b1} (MICCAI'22)& 0.6365 & 0.3087 & 0.7463  & 0.4622 & 0.2865 & 0.7331         
\\
MFH \cite{b30}   & 0.5942 & 0.3273  & 0.7541 & 0.4614 & 0.3326  & 0.7237      
\\
BlockTucker \cite{b31}  & 0.6268  & 0.2964 & 0.7631  & 0.4552 & 0.3122   & 0.7612          \\
MUTAN \cite{b32}  & 0.6298 & 0.3379 & 0.7714  & 0.4784   & 0.3244  & 0.7694       
\\
GVLE-LViT \cite{b3} (ICRA'23) & 0.6512 & 0.3365 & 0.7739  & 0.4565 & 0.2679 & 0.7296 
\\
CAT-ViL DeiT \cite{b6} (MICCAI'23)  & 0.6436 & 0.3421 & 0.7712   & 0.4765 & 0.3467   & 0.7621         \\
Surgical-VQLA++ \cite{b7} (INFORM FUSION'25) &0.6573  &0.3203  & 0.7956   & 0.4983 & 0.4365 & \textbf{0.7764}
\\
EnVR-LPKG \cite{hao2025enhancing}(JBHI'25)  & 0.6723 & 0.3826 & 0.7894 & 0.4786 & 0.4126 & 0.7438 \\
\hline \hline
Surgical-MambaLLM (our) & \textbf{0.6964} & \textbf{0.4110} & \textbf{0.8027} & \textbf{0.5191} & \textbf{0.4406} & 0.7648 \\  \hline
\end{tabular}%
}
\end{table*}

\begin{table*}[t]
\centering
\caption{Ablation study on different variants of our approach on the EndoVis-18 and EndoVis-17 datasets. Baseline, M1, M2, M3, M4, and M5 represent diverse ablation models, while Surgical-MambaLLM is our proposed model. Fusion Module represents the cross-modal fusion module in the network, Scanning Mode means different scanning modes utilized in the CBMI module}
\label{tab:2}
\resizebox{1\textwidth}{!}{%
\begin{tabular}{@{}c|c|c|ccc|ccc@{}}
\hline
\multirow{2}{*}{Models} & \multirow{2}{*}{Scanning Mode} & \multirow{2}{*}{Fusion Module}  & \multicolumn{3}{c|}{EndoVis-18} & \multicolumn{3}{c}{EndoVis-17}                     \\ \cline{4-9}
  &&& Acc & F-Score & mIoU & Acc & F-Score & mIoU            \\ \hline \hline
 
Baseline & $\times$&$\times$ & 0.6537  & 0.3595  & 0.7742  & 0.4216  & 0.3494 &0.7315     \\
M1  & Simple 1D Scan    & CBMI  &0.6644 &0.3335 &0.7951 &0.4826 & 0.3116  &0.7434  \\
M2  & Bi-Scan \cite{b34}& CBMI &0.6615 &0.3663 &0.7915 &0.4256 & 0.3774  &0.7611         \\
M3    & Cross-Scan \cite{b33} &CBMI &0.6834 &0.3420 &0.7965 & 0.4675 & 0.3669 &0.7348     \\       
M4    & SIP  &Mamba &0.6833  &0.3795 &0.7847 &0.4778  &0.4011 &0.7506\\
M5    & $\times$ &Transformer &0.6610  &0.3524 &0.7895 &0.4766  &0.3947 &0.7559\\
\hline  \hline
    Surgical-MambaLLM (our) &SIP & CBMI& \textbf{0.6964} & \textbf{0.4110} & \textbf{0.8027} & \textbf{0.5191} & \textbf{0.4406} & \textbf{0.7648} \\  \hline
\end{tabular}%
}

\end{table*}

\subsection{Implementation Details}
We employ the CLIP-ViT-B/32 \cite{b25} pre-trained model as our vision encoder. The backbone of the CBMI module is Mamba2-130M \cite{b23}, which comprises 24 layers. Our LLM is based on InternLM-7B \cite{b28}, featuring 32 layers. The projector is a 2-layer MLP. The classification head is a simple linear layer, while the location head is a 4-layer MLP. Our model is implemented using PyTorch and trained on a workstation equipped with 6 NVIDIA GeForce RTX 3090 GPUs. Following \cite{b3,b6,b7}, we use Accuracy (Acc) \cite{b13}, F-Score \cite{b13}, and mIoU \cite{b14} as evaluation metrics. The batch size per GPU is 8. We utilize the AdamW optimizer \cite{b15} with a learning rate of $1\times 10^{-5}$ for the first stage and $1\times 10^{-6}$ for the second stage, and a dropout rate of 0.1.

\subsection{Experiment Results}
\textbf{Comparisons with SOTAs.} In Table~\ref{tab:1}, we compare the performance of Surgical-MambaLLM with other state-of-the-art models on the EndoVis-18 and EndoVis-17 datasets. The comprehensive results indicate that Surgical-MambaLLM outperforms other methods on both datasets, demonstrating its superiority. Specifically, Surgical-MambaLLM achieves significantly higher Accuracy, F-Score, and mIoU on the EndoVis-18 dataset compared to other methods. This indicates that Surgical-MambaLLM exhibits better visual understanding, visual reasoning, and localization capabilities when handling complex surgical scenes. On the EndoVis-17 dataset, Surgical-MambaLLM outperforms all SOTA methods in Accuracy and F-score, although its mIoU is a little lower than Surgical-VQLA++ \cite{b7}. This suggests that while Surgical-MambaLLM maintains strong scene understanding capabilities on the EndoVis-17 dataset, there is still room for improvement in localization prediction on the external validation dataset. Overall, through quantitative evaluation, we can conclude that our Surgical-MambaLLM possesses excellent visual reasoning and localization capabilities. Despite room for improvement in a certain metric, its overall performance significantly outperforms existing Surgical-VQLA methods, validating its superiority.

\textbf{Ablation Studies.} To validate the effectiveness of the components in our proposed Surgical-MambaLLM model, we conduct ablation study experiments on the EndoVis-18 and EndoVis-17 datasets, with the results presented in Table~\ref{tab:2}. First, we evaluate the effectiveness of the CBMI module by comparing the Baseline model with our Surgical-MambaLLM model. The baseline model directly concatenates textual and visual features as input to the LLM, bypassing the CBMI module. Compared to the baseline, Surgical-MambaLLM shows significant improvements across all metrics on both datasets, confirming the efficacy of the CBMI module. 

Additionally, we explore the impact of different scanning modes on the model's performance by comparing M1, M2, and M3 with our Surgical-MambaLLM model. In M1, M2, and M3, the CBMI module employs different scanning methods: M1 employs simple 1D unidirectional scan, while M2 and M3 utilize the commonly used Bi-Scan mode \cite{b34} and Cross-Scan \cite{b33} mode. Our Surgical-MambaLLM employs the SIP scanning mode we designed. Compared to M1, M2, and M3, Surgical-MambaLLM achieves the best performance in terms of Accuracy, F-Score, and mIoU, validating the effectiveness of the SIP scanning mode. 

Finally, we validate the effectiveness of Mamba2. In M4, Mamba2 is replaced with Mamba in CBMI while maintaining the same scanning mode; in M5, we employ the widely-used Cross-Attention \cite{vaswani2017attention} for feature fusion. Experimental results demonstrate that our model achieved superior performance in terms of accuracy, F-Score, and mIoU compared to both M4 and M5, thereby proving the superiority of Mamba2. In summary, these ablation studies demonstrate the critical role of each component in enhancing the overall performance of our Surgical-MambaLLM, underscoring the benefits of the CBMI module, SIP scanning mode, and Mamba2 model.

%\textbf{Discussion and Limitations.} The experiment results demonstrate that Surgical-MambaLLM significantly outperforms existing state-of-the-art models on the EndoVis-18 and EndoVis-17 datasets. This highlights the model's exceptional capabilities in visual understanding, reasoning, and localization when handling complex surgical scenarios. Ablation studies further validate the effectiveness of the CBMI module and the SIP scanning mode, showing substantial improvements over the baseline model and other models with different scanning modes.

%However, despite these advancements, the performance of Surgical-MambaLLM on the EndoVis-17 dataset, while strong in terms of accuracy, still has room for improvement in F-Score and mIoU. This indicates that although our model performs well in scene understanding, its grounding capability on external validation datasets requires further optimization. Additionally, the complexity of multimodal data fusion and the high computational requirements of the model present challenges for practical application, necessitating further research and improvement. 

\section{Conclusion}
In this paper, we propose an innovative method, Surgical-MambaLLM, which is the first method integrating the Mamba model and LLM in Surgical-VQLA. To meet the unique characteristics of surgical scenarios, we designed the SIP scanning mode to comprehensively scan surgical images, enhancing the Mamba2 model's spatial awareness of surgical scenes. Additionally, we introduce the CBMI module to achieve multimodal fusion, thereby improving the model's spatial understanding and cross-modal fusion capabilities for surgical images. Experiment results demonstrate that our Surgical-MambaLLM model outperforms SOTA methods on the EndoVis17-VQLA and EndoVis18-VQLA datasets, significantly improving the performance of the Surgical-VQLA task. Future work will focus on improving the model's performance on external validation sets and further improving the grounding capability. We plan to expand the dataset, optimize the model architecture, and refine training strategies to better leverage the potential of LLMs in the field of surgical robotics.
\noindent\textbf{Acknowledgment.} 
This work is supported by the Guangdong Science and Technology Department (No. 2024ZDZX2004) and the Guangdong Provincial Key Lab of Integrated Communication, Sensing and Computation for Ubiquitous Internet of Things (No.2023B1212010007).

\noindent\textbf{Disclosure of Interests.} 
The authors have no competing interests to declare that are relevant to the content of this article.
%
% ---- Bibliography ----
%
% BibTeX users should specify bibliography style 'splncs04'.
% References will then be sorted and formatted in the correct style.
%
\bibliographystyle{splncs04}
\bibliography{Paper-2478}
%
% \begin{thebibliography}{8}
% \bibitem{ref_article1}
% Author, F.: Article title. Journal \textbf{2}(5), 99--110 (2016)

% \bibitem{ref_lncs1}
% Author, F., Author, S.: Title of a proceedings paper. In: Editor,
% F., Editor, S. (eds.) CONFERENCE 2016, LNCS, vol. 9999, pp. 1--13.
% Springer, Heidelberg (2016). \doi{10.10007/1234567890}                                          

% \bibitem{ref_book1}
% Author, F., Author, S., Author, T.: Book title. 2nd edn. Publisher,
% Location (1999)

% \bibitem{ref_proc1}
% Author, A.-B.: Contribution title. In: 9th International Proceedings
% on Proceedings, pp. 1--2. Publisher, Location (2010)

% \bibitem{ref_url1}
% LNCS Homepage, \url{http://www.springer.com/lncs}. Last accessed 4
% Oct 2017
% \end{thebibliography}
\end{document}